\documentclass[10pt,twocolumn,letterpaper]{article}

\usepackage{iccv}
\usepackage{times}
\usepackage{epsfig}
\usepackage{graphicx}
\usepackage{amsmath}
\usepackage{amssymb}

\usepackage{times}
\usepackage{epsfig}
\usepackage{graphicx}
\usepackage{amsmath}
\usepackage{amssymb}

\usepackage{hyperref}
\usepackage{url}
\usepackage[utf8]{inputenc} 
\usepackage[T1]{fontenc}    
\usepackage{hyperref}       
\usepackage{url}            
\usepackage{booktabs}       
\usepackage{amsfonts}       
\usepackage{nicefrac}       
\usepackage{microtype}      
\usepackage{xcolor}         

\usepackage{caption}
\usepackage{subcaption}
\usepackage{amsmath,amssymb,amsfonts}
\usepackage{algorithmic}
\usepackage{graphicx}
\usepackage{textcomp}
\usepackage{xcolor}
\usepackage{tabto}
\usepackage{comment}
\usepackage{booktabs}
\usepackage{diagbox}
\usepackage{flexisym}
\usepackage{bbding}
\usepackage{multirow}
\usepackage{makecell}
\usepackage{amsmath, bm}
\usepackage{tikz}
\usepackage{eso-pic}
\usepackage{algorithm}
\usepackage{algorithmic}
\usepackage{wrapfig}

\usepackage{bbding}

\usepackage{colortbl}

\usepackage{threeparttable}

\newcommand{\av}[0]{\ensuremath{\boldsymbol{a}} }
\newcommand{\bv}[0]{\ensuremath{\boldsymbol{b}} }
\newcommand{\cv}[0]{\ensuremath{\boldsymbol{c}} }

\newcommand{\pv}[0]{\ensuremath{\boldsymbol{p}} }

\newcommand{\xv}[0]{\ensuremath{\boldsymbol{x}} }
\newcommand{\yv}[0]{\ensuremath{\boldsymbol{y}} }

\newcommand{\piv}[0]{\ensuremath{\boldsymbol{\pi}} }

\newcommand{\phiv}[0]{\ensuremath{\boldsymbol{\phi}} }

\newcommand{\thetav}[0]{\ensuremath{\boldsymbol{\theta}} }

\newcommand{\alphav}[0]{\ensuremath{\boldsymbol{\alpha}} }

\def\bb{\textcolor{blue}}

\iccvfinalcopy 

\def\iccvPaperID{12384} 

\ificcvfinal\fi

\begin{document}

\title{Prototypes-oriented Transductive Few-shot Learning with Conditional Transport}

\author{Long Tian \textsuperscript{1}, Jingyi Feng \textsuperscript {1}, Wenchao Chen \textsuperscript {2,\Envelope}, Xiaoqiang Chai \textsuperscript{1}, Liming Wang \textsuperscript{1}, Xiyang Liu \textsuperscript{1}, Bo Chen \textsuperscript{2} \\
\textsuperscript{1} National Pilot School of Software Engineering, \quad \textsuperscript{2} National Key Laboratory of Radar Signal Processing \\
\textsuperscript{1, 2} Xidian University, Xi’an, Shanxi 710071, China \\
{\tt\small Email: \{tianlong, chenwenchao\}@xidian.edu.cn}
}

\maketitle
\ificcvfinal
\thispagestyle{empty}
\fi

\begin{abstract}
   Transductive Few-Shot Learning (TFSL) has recently attracted increasing attention since it typically outperforms its inductive peer by leveraging statistics of query samples.
   However, previous TFSL methods usually encode uniform prior that all the classes within query samples are equally likely, which is biased in imbalanced TFSL and causes severe performance degradation.
   Given this pivotal issue, in this work, we propose a novel Conditional Transport (CT) based imbalanced TFSL model called {\textbf P}rototypes-oriented {\textbf U}nbiased {\textbf T}ransfer {\textbf M}odel (PUTM) to fully exploit unbiased statistics of imbalanced query samples, which employs forward and backward navigators as transport matrices to balance the prior of query samples per class between uniform and adaptive data-driven distributions. For efficiently transferring statistics learned by CT, we further derive a closed form solution to refine prototypes based on MAP given the learned navigators. The above two steps of discovering and transferring unbiased statistics follow an iterative manner, formulating our EM-based solver.
   Experimental results on four standard benchmarks including miniImageNet, tieredImageNet, CUB, and CIFAR-FS demonstrate superiority of our model in class-imbalanced generalization.
\end{abstract}

\section{Introduction}
Deep learning based methods have gained a great success in many real-world applications, such as image classification \cite{he2016deep,simonyan2014very,zagoruyko2016wide},  natural language processing \cite{chen2020bidirectional,wang2020deep,zhang2018whai}, and so on, thanks to its powerful nonlinear representation ability \cite{lu2021learning,scarselli1998universal}. However, their outstanding performances heavily rely on large-scale annotated training data \cite{deng2009imagenet,lin2014microsoft}. 
When it comes to high cost of collecting and annotating a large amount of data, a major research effort is being dedicated to low data regimes, in which overfitting occurs in optimizing deep learning architectures. Few-shot Learning (FSL) tackles these challenges and has triggered substantial interests within the community \cite{fei2006one,miller2000learning}, it aims to train a model from the base classes so as to generalize on the tasks sampled from the novel classes that were never seen during training.
There are two major lines of works from the perspective of inference paradigms, namely inductive methods \cite{snell2017prototypical,chen2019closer} and their transductive counterparts \cite{dhillon2019baseline,ren2018meta}. Both of them adopt well trained models for evaluating new tasks sampled from the novel classes. The former ones only utilize support samples in a supervised way while the later ones leverage both support and query samples in a semi-supervised manner.


\begin{figure}[!t]
\centerline{\includegraphics[width=1.0\linewidth,height=0.5\linewidth]{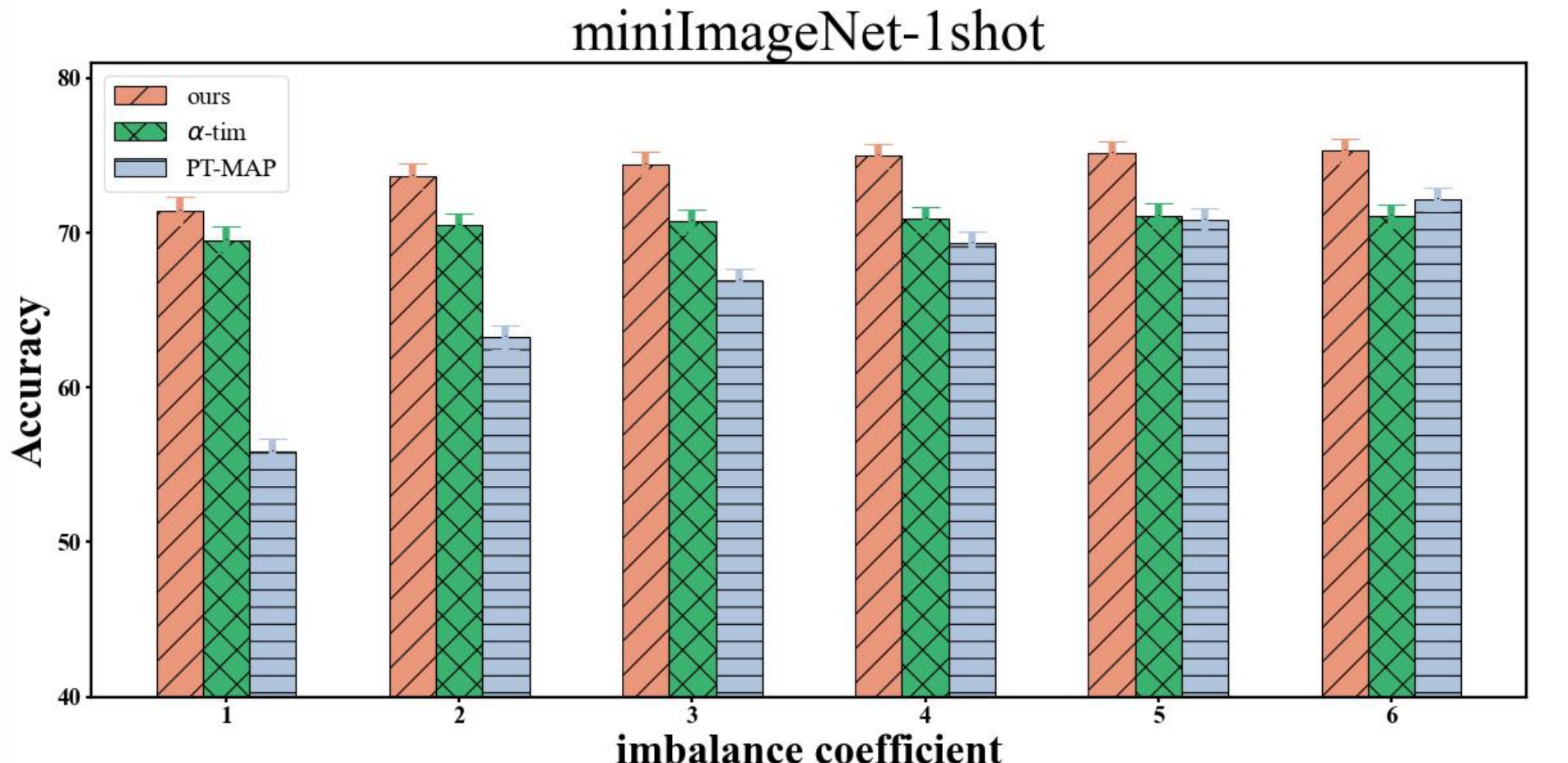}}
\caption{5-way 1-shot image classification results with miniImageNet \cite{ravi2017optimization} dataset on 3000 tasks. Our model achieves the best accuracy in a wide range of imbalanced coefficients compared with PT-MAP \cite{hu2021leveraging} and $\alpha$-TIM \cite{veilleux2021realistic}.}
\label{fig:motivation}
\end{figure}

Transductive inference is widely used in FSL since it can sufficiently discover statistics of query samples compared with its inductive peer. As a representative example, Hu \etal \cite{hu2021leveraging} propose PT-MAP for first employing Wide ResNet (WRN) \cite{zagoruyko2016wide} trained with S2M2 \cite{mangla2020charting} on base classes to extract features of new tasks, and then adjusting prototypes from novel classes by aggregating similar query samples per class with the help of transport matrix derived from the Sinkhorn algorithm \cite{cuturi2013sinkhorn}. Setting marginal distributions of prototypes as uniform 
encodes strong prior that all the classes within query samples are equally likely, although, it works pretty well in perfectly class-balanced query samples 
. However, it is proven to be detrimental in class-imbalanced applications where query samples come with arbitrary and unknown label marginals \cite{ochal2021few}, and classification accuracy drops dramatically, as shown in Fig. \ref{fig:motivation}. To narrow this gap, Veilleux \etal \cite{veilleux2021realistic} propose 
$\alpha$-TIM for handling class-distribution variations effectively.
Unfortunately, we have observed the phenomena as shown in Fig. \ref{fig:motivation} that $\alpha$-TIM can not achieve consistent improvement in various class-imbalanced levels, thus restricting its real-world applications.
Most recently, other more complicated methods are also be arisen \cite{lazarou2021iterative,hu2023adaptive}. Hu \etal \cite{hu2023adaptive} propose a clustering method for capturing the class-wise distributions properly.

To this end, for class-imbalanced TFSL, we develop a novel and simple Prototypes-oriented Unbiased Transfer Model (PUTM). It firstly constructs transferable class-wise prototypes iteratively for capturing the unbiased statistics within class-imbalanced query samples.
Then it performs few-shot classification via the similarities between prototypes and query samples when the transfer arrives at convergence. The key to obtain unbiased transferable statistics is estimating class-wise marginal distributions precisely \cite{lazarou2021iterative}.
To achieve this, instead of utilizing doubly stochastic transport matrix constrained with fixed uniform-distributed marginal prior between prototypes and query samples as PT-MAP does, we employ task-adaptive forward and backward transport matrices with learnable marginal priors constrained by minimizing Conditional Transport (CT) distance \cite{zheng2021exploiting} between prototypes and query samples. CT is a tool for measuring the cost in transporting the mass in one distribution to match another in an exchangeable but asymmetrical manner given a specific point-to-point cost function and an optimizable forward and backward transport matrices, also called navigators in the CT scope.
To better calibrate prototypes with the help of learned transport matrices and query samples, we further derive a closed-form solution to refine prototypes based on MAP in an iterative manner. Decisions are finally made using the combination of forward and backward transport matrices when the iteration converges.
Experiments on four benchmarks demonstrate that our proposed PUTM can learn task-adaptive prior in various class-imbalanced coefficients, paving a new way to transfer the unbiased statistics of query samples to prototypes.

Our contributions can be summarized as follows:
\begin{itemize}
    \item  We find that the performance of class-imbalanced FSL heavily relies on the accurate estimation of class-wise marginal distribution. Meanwhile, we are the first one to find that there exists a nature connection between CT theory and such marginal distribution estimation. 
    \item We propose a novel and simple PUTM for class-imbalanced FSL based on connections mentioned above. It firstly embeds CT distance to measure the differences between prototypes and query samples for achieving better task-adaptive unbiased statistic migration elegantly. Then it calibrates prototypes using the transferred statistics in an iterative manner.
    \item We provide extensive experiments and comparisons on four standard benchmarks for verifying that our model achieves competitive results on both class-balanced and class-imbalanced (in a wide range of imbalanced coefficients) FSL settings.
\end{itemize}
\section{Task Formulation}
\label{dirichlet}
We divide the whole annotated dataset into two disjoint subsets: a base set $\mathcal{D}_{base}$ with $\mathcal{B}$ classes and a novel set $\mathcal{D}_{novel}$ with $\mathcal{N}$ classes. To build a common N-way K-shot task \cite{vinyals2016matching,snell2017prototypical}, we first randomly sample N classes from $\mathcal{N}$ novel classes, and then pick K samples per class for the support set $\mathcal{S}=\{\xv_{i,j}^s, \yv_{i,j}^s\}_{i=1,j=1}^{N,K}$ to train or fine-tune the model. Finally, $K^{\prime}_i$ query samples being non-overlapped with support samples per class are randomly picked from the $N$ support classes for evaluating the performance of the model, formulating query set as $\mathcal{Q}=\{\xv_{i,j}^q,\yv_{i,j}^q\}_{i=1,j=1}^{N,K^{\prime}_i}$. For simplicity, we set $M=\sum_{i=1}^N K_i^{\prime}$.

\begin{figure*}[!t]
\centerline{\includegraphics[width=1\linewidth,height=0.3\linewidth]{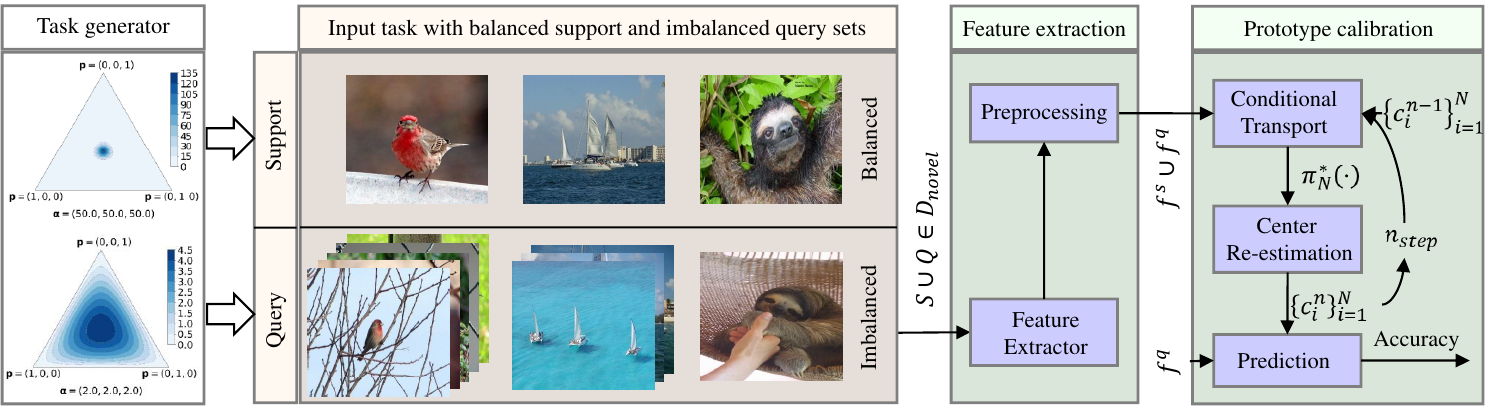}}
\caption{Illustration of the our model. Tasks are generated from Dirichlet-distributed generator parameterized by $\av$. We first adopt a well trained feature extractor \cite{mangla2020charting} and a channel-wise feature adaptation \cite{yang2021free} to obtain the input features. Transductive inference is then conducted on the pre-processed features by: i) optimizing the transport matrices using CT; ii) refining the class-wise prototypes via MAP statistics aggregation; iii) making predictions after the above two steps iterate until convergence.}
\label{fig:whole framework}
\end{figure*}

\textbf{Class-imbalanced TFSL:}
Veilleux \etal \cite{veilleux2021realistic} figure out that the number of query samples per class should be arbitrary and unknown instead of being strictly equal. That is $K^{\prime}$ is a function of class and changes with class indexes. To describe such behavior, they develop a class marginal with Dirichlet distribution for generating query samples. Specifically, the class-wise marginal distribution of query samples per task $\pv$ are determined by:
\begin{equation} \label{eq-1}
    f_{Dir}(\pv; \av) = \frac{1}{B(\alphav)} \prod_{i=1}^{N} p_i^{\alpha_i-1}
\end{equation}
where $\pv=(p_1,...,p_N)$ stands for class-wise distribution, $\alphav=(\alpha_1,...,\alpha_{N})$ denotes Dirichlet-distributed parameter controlling imbalanced ratio of query samples, $B(\av)=\frac{\prod_{i=1}^{N} \Gamma(a_i)}{\Gamma(\sum_{i=1}^{N}a_i)}$ represents multivariate Beta function, and $\Gamma(\cdot)$ is Gamma function.
Smaller $\av$ means higher imbalanced level, and vice versa. 
Taking $\alphav=2$ as an example, total number of query samples is 75, then the number of query samples per class could be distributed as $K^{\prime}=[10, 5, 2, 35, 18]$, which is more realistic compared with the class-balanced situation whose $K^{\prime}_i=15,i=1,...,N$. Besides, it also poses some difficulties such as overfitting to the classes whose query samples are extremely few.


\section{Our Model}
\subsection{Overall Method}
In this work, we propose a novel model called PUTM for class-imbalanced transductive few-shot image classification. 
Considering the correlations between support and query samples, it is reasonable to refine the representations of class-wise prototypes initialized by support samples with the statistics of query samples.
At this moment, the key is how to transfer the unbiased statistics from query samples to prototypes for achieving the best calibration. Moreover, the class-imbalanced setting is more challenging due to its unknown and arbitrary prior distribution of query samples. 
To address this challenge, we pioneered to depict the important but unknown prior with well-defined CT theory. Specifically, the prior could be measured by the transport mass from query samples to each prototype by summing the combination of forward and backward transport matrices optimized by CT loss along every query sample. Given the well-optimized prior, a prototype refinement algorithm is easy to be derived according to the MAP criterion. Considering the fact that the discriminative prototypes and well-optimized prior are coupled, we further employ an EM-solver to transfer the unbiased statistics efficiently and precisely.
The overview of the proposed method can be found in Fig. \ref{fig:whole framework}.


\subsection{CT Theory}
\label{CT}

CT \cite{zheng2021exploiting} is a powerful tool developed for measuring distance between any two probability distributions most recently, the effectiveness of which is mostly verified on deep generative models \cite{karras2020analyzing}, multi-label classification \cite{li2023patchct}, and pre-training on big models \cite{tanwisuth2023pouf}. In this work, we expand the territory of CT for discrete distributions under class-imbalanced TFSL scenario.
Denote $p(x)=\sum_{i=1}^n a_i \delta_{x_i}$ and $q(y)=\sum_{j=1}^m b_j \delta_{y_j}$ as source and target discrete distributions with $n$ and $m$ points, respectively. In this case, $\av \in \Delta^n$ and $\bv \in \Delta^m$, where $\Delta^n$ and $\Delta^m$ separately denote the probability simplex of $\mathbb{R}^n$ and $\mathbb{R}^m$.
The CT distance between $p(x)$ and $q(y)$ can be expressed as a combination of forward CT $\mathcal{C}_{\phiv}(x \rightarrow y)$ and a backward CT $\mathcal{C}_{\phiv}(x \leftarrow y)$:
\begin{equation} \label{eq-2}
\small
    \mathcal{C}_{\phiv, \rho}(p, q) = \min_{\phiv} \sum_{x \in p, y \in q} \rho \mathcal{C}_{\phiv}(x \rightarrow y) + (1 - \rho) \mathcal{C}_{\phiv}(x \leftarrow y)
\end{equation}
where $\phiv$ is parameters of CT, $\rho$ is the trade-off factor controlling the strengths between the forward and backward CTs.
To better understand how the CT works when minimizing Eq. \ref{eq-2}, we take the forward CT as an example:
\begin{equation} \label{eq-3}
\small
\begin{aligned}
    \mathcal{C}_{\phiv}(x \rightarrow y) & = \mathbb{E}_{y_{1:m} \sim q} \mathbb{E}_{x \sim p} [\sum_{j=1}^{m} c(x, y_j) \hat{\pi}_{m}(y_j|x, \phiv)]
\end{aligned}
\end{equation}
where $c(\cdot, \cdot)$ could be some well defined distance functions, such as Euclidean or Cosine, $\hat{\pi}_{m}(y_j|x, \phiv)$ is forward transport matrix, also called forward navigator in the scope of CT, satisfying $\sum_{j=1}^m \hat{\pi}_{m}(y_j|x, \phiv)=1$, it describes the possibility of transporting $x$ to $y_j$ and can be formulated as:
\begin{equation} \label{eq-4}
\small
    \hat{\pi}_{m}(y_j|x, \phiv)=\frac{e^{-d_{\phiv}(x,y_j)}}{\sum_{j^{\prime}}^{m} e^{-d_{\phiv}(x,y_{j^{\prime}})}}
\end{equation}
where $d_{\phiv}(\cdot, \cdot)$ represents measurement function, such as MLP implemented by Zheng \etal \cite{zheng2021exploiting}. Therefore, the forward CT can be interpreted as the expected cost of following the forward navigator to stochastically transport a random source point $x$ to one of the $m$ randomly instantiated "anchors" of the target distribution. By minimizing Eq. \ref{eq-3}, an unbiased statistical estimation of the two distributions arrives. Proofs can be found in Zheng \etal \cite{zheng2021exploiting}. Hence, it is reasonable to transfer statistics of one distribution to another.

\subsection{PUTM}
\label{use CT}
Our proposed PUTM aims to exploit discriminative statistics from query samples for reducing the gap between the estimated prototypes and their ground truth prototypes. However, the ground truth prototypes are always intractable in practice. A reasonable approach to obtain unbiased prototypes is to aggregate discriminative statistics from both support and query samples following a semi-supervised manner \cite{zhang2021sill}. 
Different from the standard semi-supervised problems, the query samples are usually limited per class, especially in class-imbalanced case. 
Follow up, we present the importance of unbiased transfer in semi-supervised FSL setting, the algorithm of realizing unbiased exploitation with CT and transductive inference with unbiased transfer in turn.

\subsubsection{Unbiased Statistics Transfer Matters in TFSL}
\label{USM}
Given a few-shot task, unbiased prototypes arrives if and only if the aggregated query samples are noiseless, which should heavily rely on the unbiased statistics transfer.
Leveraging similarities between query samples and prototypes is a common way to realize statistics transfer in transductive few shot image classification. 
Given the transport matrix $\piv^n \in \mathbb{R}^{M \times N}$ at the $n$-th iteration, it measures the closeness of prototypes $\{\cv_i^n\}_{i=1}^N$ and query samples $\{\xv_{i,j}^q\}_{j=1}^{K_i^{\prime}}$. For simplicity, we rewrite all query samples as $\{\xv_{r}^q\}_{r=1}^M$. Thus, the refined prototypes $\{\cv_i^{n+1}\}_{i=1}^{N}$ can be expressed as:
\begin{equation} \label{eq-5}
\small
    \cv_i^{n+1} = \frac{\sum_{r=1}^{M} \piv_{r,i}^n \xv_{r}^q + \sum_{j=1}^{K} \xv_{i,j}^s}{\sum_{r^{\prime}=1}^{M} \piv_{r^{\prime},i}^n+K},\quad i=1,...,N
\end{equation}
where $\piv_{r,i}^n=\frac{e^{-d_{\phiv}(\xv_r^q, \cv_i^n)}}{\sum_{r^{\prime}=1}^{M}e^{-d_{\phiv}(\xv_{r^{\prime}}^q, \cv_i^n)}}$, $d_{\phiv}(\cdot, \cdot)$ denotes the well defined measurement function, and we denote optimal transport matrix as $\piv^*$. To guarantee the unbiased refinement of prototypes, $\piv^*$ need to be precise, which is usually difficult to be satisfied in practice. Fortunately, partial noisy is tolerable. For example, given 15 query samples per class, a few wrong transferring such as 1 or 2 samples is tolerable in calibrating prototypes. As for class-imbalanced case, a tiny mistake may result in disastrous consequence on prototypes calibration. For the classes having only 1 or 2 query samples, the tolerances of error transmissions should be extremely low.
Therefore, to improve the robustness of calibrated prototypes in class-imbalanced TFSL, transferring unbiased statistics of query samples with $\piv^*$ becomes crucial.

There are mainly two lines of works in defining $\piv^*$: i) Liu \etal \cite{liu2020prototype} employ Cosine function \cite{liu2020prototype}; ii) Hu \etal \cite{hu2021leveraging} use Optimal Transport (OT). Cosine function may not powerful enough to fully reflect the relationships between prototypes and query samples since it ignores various importance between each two points.
OT describes a doubly stochastic transport matrix such that $\Pi(p, q)=\{\piv^*|\sum_{r=1}^M \piv^*_{r,i}=a_i, \sum_{i=1}^N \piv^*_{r,i}=b_r\}$. Unfortunately, we usually do not have any priors on distributions of $p$ and $q$, thus, we usually have $a_i=\frac{1}{N},i=1,...,N$ and $b_r=\frac{1}{M},r=1,...,M$, which should be contradictory to class-imbalanced setting. For example, once we employ Dirichlet distribution with $\alphav=\textbf{2}_N$ to mimic the generation of query samples, $a_i=\frac{1}{N}$ will never fit the data.
According to the analysis of CT, it replaces the uniform constraint on prior distribution of classes derived from query samples with a data-driven manner, thus ensuring to transfer the unbiased statistics more flexible and efficient.

\subsubsection{Unbiased Statistics Exploitation with CT}
\label{UST}
As we can see, CT has an appealing property of exploiting unbiased statistics thanks to its task-adaptive prior rather than uniform-distributed one as OT. Here after, we present details of achieving such unbiased exploitation with CT.

As shown in Fig. \ref{fig:whole framework}, given a class-imbalanced few-shot task in Sec. \ref{dirichlet}, we first employ a well-trained feature extractor $f_{\thetav^*}(\cdot)$ to get features of support and query samples as $f^s_{i,j}=f_{\thetav^*}(\xv_{i,j}^s)$ and $f^q_r=f_{\thetav^*}(\xv_{r}^q)$, where $\thetav^*$ is the set of optimized parameters, being fixed after training on the base classes following Mangla \etal \cite{mangla2020charting}. Then, we conduct channel attention with preprocessing developed by Yang \etal \cite{yang2021free}. Finally, unbiased statistics exploitation is realized by optimizing transport matrices constrained by CT theory. 

Specifically, the initial prototypes are averaged on the features of support set as $\cv_i^0 = \frac{1}{K} \sum_{j=1}^K f_{i,j}^s$, $i=1,...,N$,
where the superscript of $\{\cv_i^0\}_{i=1}^N$ describes the generation of prototypes is zero. They will be updated subsequently, detailed in Sec. \ref{tiwp}.
At this moment, we treat $\{\cv_i^0\}_{i=1}^{N}$ and $\{f_{r}^q\}_{r=1}^M$ as samples from the two discrete distributions, and utilizing CT to learn the transport matrices for exploiting unbiased statistics. According to the analysis of Eq. \ref{eq-2}, an unbiased sample estimate of CT given query features $\{f_{r}^q\}_{r=1}^M$ and prototypes $\{\cv_i^0\}_{i=1}^{N}$ can be expressed as:
\begin{equation} \label{eq-ctloss}
    \begin{split}
    & \mathcal{L}_{\phiv,\thetav^*,\rho}(f_{1:M}^q, \cv^0_{1:N}) = \sum_{i=1}^{N} \sum_{r=1}^{M} c(\cv_i^0, f_r^q) \cdot \\
    & \quad \quad \quad \quad \,(\frac{\rho}{M} \hat{\piv}_N(\cv_i^0|f_r^q,\phiv) + \frac{1-\rho}{N}\hat{\piv}_M(f_r^q|\cv_i^0,\phiv))
    \end{split}
\end{equation}
where $\hat{\piv}_N$ and $\hat{\piv}_M$ separately denote the forward and backward transport matrices. Again substituting query features $\{f_{r}^q\}_{r=1}^M$ and prototypes $\{\cv_i^0\}_{i=1}^{N}$ into Eq. \ref{eq-4}, we have:
\begin{align}
    & \hat{\piv}_{N}(\cv_i^0|f_r^q, \phiv)=\frac{e^{-d_{\phiv}(f_r^q,\cv_i^0)}}{\sum_{i^{\prime}=1}^{N} e^{-d_{\phiv}(f_r^q,\cv_{i^{\prime}}^0)}} \\
    & \hat{\piv}_{M}(f_r^q|\cv_i^0, \phiv)=\frac{e^{-d_{\phiv}(f_r^q,\cv_i^0)}}{\sum_{r^{\prime}=1}^{M} e^{-d_{\phiv}(f_{r^{\prime}}^q,\cv_i^0)}}
\end{align}
The forward transport matrix represents the probability of query samples belonging to the corresponding support classes.
The two transport matrices with the optimal parameter $\phiv^*$ can be obtained by minimizing the approximated CT estimation of Eq. \ref{eq-ctloss}. Once $\hat{\piv}_{N}^*$ and $\hat{\piv}_{M}^*$ with the optimal $\phiv^*$ are obtained, the transport matrix described in Sec. \ref{USM} for unbiased statistics exploitation can be naturally realized. To better understand this, we depict the mechanisms of the transport matrices developed above from three perspectives: 1) the distance function in transport matrices is implemented with DNNs, hence, they are able to learn a comprehensive distance metric beyond a simple Cosine function; 2) the CT distance is proven to be unbiased given samples of two distributions $p$ and $q$ in \cite{zheng2021exploiting}; 3) The forward transport matrix holds free for sum along query samples given a specific prototype, namely, $\sum_{r=1}^{M}\hat{\piv}_{N}(\cv_i^0|f_r^q, \phiv^*)=C$, and $C$ is not restricted to be equal for each class, thus embedding task-adaptive prior instead of uniform distribution as PT-MAP \cite{hu2021leveraging} does. 
On the other hand, the backward transport matrix holds tight for sum along query samples given a specific prototype, that is, $\sum_{r=1}^{M}\hat{\pi}_{M}(f_r^q|\cv_i^0, \phiv^*)=1$, hence, uniform-distributed prior holds, and it is proven to be underfitting in class-imbalanced case \cite{veilleux2021realistic}.
In practice, we employ a trade-off to balance the strengths of the two transport matrices by summing them together with a soft coefficient $\rho$. For simplicity, we only use the forward transport matrix in the rest of the work.

\begin{algorithm} [t]
 \caption{Work flow of PUTM for one task.}
 \small
\label{algorithm1}
\begin{algorithmic}

\STATE \textbf{Input:} Support samples $\{\xv_{i,j}^s\}_{i,j=1}^{N,K}$ and class-imbalanced query samples $\{\xv_r^q\}_{r=1}^M$ generated from Dirichlet distribution with parameter $\alphav$, a well-trained feature extractor $f_{\thetav^*}(\cdot)$;

\STATE \textbf{Parameters:} Distance measurement function $d_{\phiv}(\cdot, \cdot)$;

\STATE \textbf{Initialization} Hyper-parameters $\alpha$, $\rho$, $\lambda$, and learning schedule for updating $\phiv$;
Initialize prototypes $\cv_i^0 = \frac{1}{K} \sum_{j=1}^K f_{i,j}^s$, $i=1,...,N$;

\FOR{n=1: $n_{steps}$}
 
\STATE Initialize parameters $\phiv$;

\STATE $\phiv^{*} \gets argmin_{\phiv} \mathcal{L}_{\phiv,\thetav^*,\rho}(f_{1:M}^q, \cv_{1:N}^n)$;

\STATE Refining prototypes $\cv_{1:N}^{n+1}$ using Eq. \ref{eq-12} based on $\cv_{1:N}^n$;

\ENDFOR

\STATE Evaluating performance using Eq. \ref{eq-13};

\STATE \textbf{Output} Parameters $\{\phiv^*, \cv_{1:N}^{n_{step}}\}$.
\end{algorithmic}
\end{algorithm}

\subsubsection{Unbiased Statistics Transfer with EM-Solver}
\label{tiwp}
Once the forward transport matrix $\hat{\piv}_N^*$ is obtained by minimizing Eq. \ref{eq-ctloss}, it is straightforward to derive the refined prototypes by substituting $\hat{\piv}_N^*$ into Eq. \ref{eq-5}:
\begin{equation} \label{eq-10}
    \cv_i^{\prime} = \frac{\sum_{r=1}^{M} \hat{\piv}_{N,r,i}^* f_r^q+\sum_{j=1}^{K}f_{i,j}^s} {K+\sum_{r=1}^{M}\hat{\pi}_{N,r,i}^*}, \quad i=1,...,N
\end{equation}
It is worth noting that once the optimal parameters $\phiv^*$ are arrived, the only parameters left for optimizing are prototypes. Recalling our objective in Eq. \ref{eq-ctloss}, it is easy to find that the optimal prototypes at this moment could also be derived by $\frac{\partial \mathcal{L}_{\phiv^*,\thetav^*,\rho}}{\partial \cv_{1:N}^0}$, thereby Eq. \ref{eq-10} could be derived.
Instead of directly calibrating class-wise prototypes, We employ an inertia parameter $\alpha$ to update prototypes proportionally:
\begin{equation} \label{eq-11}
    \cv_i^{1} = \cv_i^{0}+\alpha(\cv_i^{\prime}-\cv_{i}^{0}), \quad i=1,...,N
\end{equation}
Right now, recursive expression emerges for updating prototypes according to Eq. \ref{eq-11}:
\begin{equation} \label{eq-12}
    \cv_i^{n+1} = \cv_i^{n}+\alpha(\cv_i^{\prime}-\cv_{i}^{n}), \quad i=1,...,N
\end{equation}
As we can see: i) updating the forward transport matrix $\hat{\piv}_N^*$ and prototypes $\cv_{1:N}^n$ are coupled; ii) $\hat{\piv}_N^*$ is a random variable determined by $\phiv$ while $\cv_{1:N}^n$ is an unknown but deterministic variable. These two facts meet conditions of implementing EM optimization, and we formulate our EM-Solver as follows: i) E Step, updating $\hat{\piv}_N^*$ by minimizing Eq. \ref{eq-ctloss} while keeping $\cv_{1:N}^n$ fixed; ii) M Step, updating $\cv_{1:N}^n$ by minimizing Eq. \ref{eq-ctloss} while keeping $\hat{\piv}_N^*$ fixed. Besides, it is worth noting that at each iteration, $\phiv^*$ need to be re-initialized. The pseudo code is provided in Alg. \ref{algorithm1}.

After $n_{step}$ loops between estimating $\hat{\piv}_N^*$ and refining $\cv_{1:N}^n$, the decision per sample is finally made by:
\begin{equation} \label{eq-13}
    y_r^q = \max_{i\in[1:N]} \hat{\piv}_N(\cv_i^{n_{step}}|f_r^q,\phiv^*), \quad r=1,...,M
\end{equation}
Statistic results of average accuracy are usually conducted using query samples on various of different tasks.

\subsection{Complexity Analysis}
$\phiv$ is implemented with MLP in our work. Taking a two-layers MLP for example, namely, $d_f \rightarrow d_1^{\phiv} \rightarrow d_2^{\phiv}$. For $T$ tasks, computational complexity originally takes $\mathcal{O}(T \times max\{NM\times d_f \times d_1^{\phiv}, NM\times d_1^{\phiv} \times d_2^{\phiv}\})$. In practice, $T$ is usually set to be large enough for reliable results, thus resulting in time-consuming evaluation of our model. To mitigate the real-time issue, we trade space for time by paralleling our proposed PUTM among all tasks. Therefore, the computational complexity is reduced to $\mathcal{O}(max\{NM\times d_f \times d_1^{\phiv}, NM\times d_1^{\phiv} \times d_2^{\phiv}\})$. Besides, we can further control the size of MLP, that is $\{d_i^{\phiv}\}_{i=1}^2$, to achieve a trade-off between accuracy and real-time efficiency.

\begin{table*}[!t]
\centering
\caption{Classification accuracy ($\%$) of SOTA methods on class-imbalanced and class-balanced settings with different datasets and backbones. 
For imbalanced case, query samples are generated following a Dirichlet distribution with $\alphav=\textbf{2}_N$. Results are averaged over 3000 tasks with 95$\%$ confidence intervals.  
Results with superscript $\dagger$ are our re-implementations based on the code provided by Veilleux \etal \cite{veilleux2021realistic}.
The best and the second best scores are marked in bold and blue, respectively.
}
\label{Table 1}
\setlength{\tabcolsep}{2mm}
\resizebox{1\hsize}{!}{
\begin{tabular}{ccccccccccccccc}
\toprule 
{\multirow{2}*{Setting}} & \multicolumn{2}{c}{\multirow{2}*{Method}} &\multicolumn{3}{c}{mini-ImageNet \quad (WRN/WRN$^{\S}$)}&\multicolumn{3}{c}{CUB \quad (RN18/WRN$^{\S}$)}&\multicolumn{3}{c}{tiered-ImageNet \quad (WRN/WRN$^{\S}$)}&\multicolumn{3}{c}{CIFAR-FS \quad (WRN$^{\S}$)}\\
\cmidrule{4-15}
&\multicolumn{2}{c}{ }  & \multicolumn{1}{c}{1-shot} & \multicolumn{1}{c}{3-shot}& \multicolumn{1}{c}{5-shot}& \multicolumn{1}{c}{1-shot} & \multicolumn{1}{c}{3-shot }& \multicolumn{1}{c}{5-shot }& \multicolumn{1}{c}{1-shot } & \multicolumn{1}{c}{3-shot}& \multicolumn{1}{c}{5-shot }& \multicolumn{1}{c}{1-shot } & \multicolumn{1}{c}{3-shot }& \multicolumn{1}{c}{5-shot } \\
\midrule
{\multirow{5}*{ Induct. ($\%$)}} 
&\multirow{5}{*}{\rotatebox{90}{\makecell{Imbalanced \\ \& Balanced}}}
& Baseline (ICLR'19)& {62.2/62.8$^{\dagger}$} & {76.9$^{\dagger}$/77.3$^{\dagger}$}& {81.9/82.9$^{\dagger}$}&{64.6/70.7$^{\dagger}$} & {82.3$^{\dagger}$/87.3$^{\dagger}$}& {86.9/90.0$^{\dagger}$}&{64.6/63.6$^{\dagger}$} & {80.0$^{\dagger}$/83.0$^{\dagger}$}& {84.9/87.2$^{\dagger}$}&{69.5$^{\dagger}$} & {83.9$^{\dagger}$}& {86.7$^{\dagger}$}\\
&& Baseline++ (ICLR'19) & {64.5/64.8$^{\dagger}$} & {77.5$^{\dagger}$/78.4$^{\dagger}$}& {82.1/82.7$^{\dagger}$}&{69.4/81.7$^{\dagger}$} & {83.3$^{\dagger}$/89.4$^{\dagger}$}& {87.1/91.2$^{\dagger}$}&{68.7/66.9$^{\dagger}$} & {80.9$^{\dagger}$/78.7$^{\dagger}$}& {85.4/82.2$^{\dagger}$}&{74.9$^{\dagger}$} & {85.0$^{\dagger}$}& {87.4$^{\dagger}$}\\
&& ProtoNet (NeurIPS'17) & {61.2/62.5$^{\dagger}$} & {76.9$^{\dagger}$/77.2$^{\dagger}$}& {81.1/81.4$^{\dagger}$}&{63.1$^{\dagger}$/69.6$^{\dagger}$} & {82.3$^{\dagger}$/85.6$^{\dagger}$}& {85.4$^{\dagger}$/88.8$^{\dagger}$}&{63.5$^{\dagger}$/64.2$^{\dagger}$} & {79.9$^{\dagger}$/80.1$^{\dagger}$}& {84.1$^{\dagger}$/84.0$^{\dagger}$}&{69.5$^{\dagger}$} & {83.8$^{\dagger}$}& {86.7$^{\dagger}$}\\
&& Simpleshot (ARXIV) & {66.2/67.1$^{\dagger}$} & {77.6$^{\dagger}$/78.2$^{\dagger}$}& {82.4/83.3$^{\dagger}$} &{70.6/81.7$^{\dagger}$} & {83.6$^{\dagger}$/89.5$^{\dagger}$}& {87.5/91.4$^{\dagger}$} &{70.7/71.4$^{\dagger}$} & {81.3$^{\dagger}$/84.2$^{\dagger}$}& {85.9/87.5$^{\dagger}$} &{75.2$^{\dagger}$} & {84.7$^{\dagger}$}& {87.4$^{\dagger}$}\\
&& HOT (NeurIPS'22) & {66.3$^{\dagger}$/69.1} & {78.5$^{\dagger}$/79.8$^{\dagger}$}& {82.8$^{\dagger}$/84.4}&{74.5$^{\dagger}$/81.2} & {85.4$^{\dagger}$/88.2$^{\dagger}$}& {87.8$^{\dagger}$/91.5}&{74.1$^{\dagger}$/75.9} & {82.6$^{\dagger}$/83.9$^{\dagger}$}& {85.4$^{\dagger}$/87.3}&{75.4} & {85.3$^{\dagger}$}& {87.5}\\
\midrule
{\multirow{16}*{ Transduct. ($\%$) }} 
&\multirow{8}{*}{\rotatebox{90}{Imbalanced}}
& PT-MAP (ARXIV) & {60.6/63.2$^{\dagger}$} & {65.9$^{\dagger}$/68.0$^{\dagger}$}& {66.8/69.3$^{\dagger}$}&{65.1/69.8$^{\dagger}$} & {71.2$^{\dagger}$/73.2$^{\dagger}$}& {71.3/74.3$^{\dagger}$}&{65.1/67.8$^{\dagger}$} & {69.1$^{\dagger}$/72.0$^{\dagger}$}& {71.0/73.2$^{\dagger}$}&{65.9$^{\dagger}$} & {69.3$^{\dagger}$}& {70.3$^{\dagger}$}\\
&& LaplacianShot (ICML'20) & {68.1/69.8$^{\dagger}$} & {78.8$^{\dagger}$/79.2$^{\dagger}$}& {83.2/83.6$^{\dagger}$}&{73.7/84.2$^{\dagger}$} & {85.7$^{\dagger}$/90.4$^{\dagger}$}& {87.7/91.4$^{\dagger}$}&{73.5/72.6$^{\dagger}$} & {82.6$^{\dagger}$/85.0$^{\dagger}$}& {86.8/87.7$^{\dagger}$}&{78.3$^{\dagger}$} & {86.1$^{\dagger}$}& {87.3$^{\dagger}$}\\
&& TIM (NeuralIPS'20) & {69.8/70.9$^{\dagger}$} & {79.7$^{\dagger}$/80.7$^{\dagger}$}& {81.6/83.2$^{\dagger}$}&{74.8/82.8$^{\dagger}$} & {84.8$^{\dagger}$/89.1$^{\dagger}$}& {86.9/90.5$^{\dagger}$}&{75.8/77.8$^{\dagger}$} &{82.5$^{\dagger}$/86.3$^{\dagger}$}& {85.4/88.3$^{\dagger}$} &{77.0$^{\dagger}$} & {84.6$^{\dagger}$}& {86.2$^{\dagger}$}\\
&& BD-CSPN (ECCV'20) & {70.4/71.2$^{\dagger}$} & {77.1$^{\dagger}$/78.6$^{\dagger}$}& {82.3/83.4$^{\dagger}$}&{74.5/85.1$^{\dagger}$} & {85.7$^{\dagger}$/90.5$^{\dagger}$}& {87.1/91.5$^{\dagger}$}&{75.4/76.3$^{\dagger}$} & {82.5$^{\dagger}$/85.0$^{\dagger}$}& {85.9/87.2$^{\dagger}$}&{79.1$^{\dagger}$} & {86.0$^{\dagger}$}& {87.3$^{\dagger}$}\\
&& $\alpha$-TIM (NeuralIPS'2021) & {69.8/70.9$^{\dagger}$} & {80.4$^{\dagger}$/81.1$^{\dagger}$}& {\textbf{\bb{84.8}}/84.9$^{\dagger}$}&{75.7/83.6$^{\dagger}$} &{87.3$^{\dagger}$/\textbf{\bb{91.2}}$^{\dagger}$}& {\textbf{\bb{89.8}}/\textbf{92.8}$^{\dagger}$}&{76.0/77.7$^{\dagger}$} & {\textbf{\bb{83.6}}$^{\dagger}$/86.6$^{\dagger}$}& {\textbf{87.8}/\textbf{\bb{89.9}}$^{\dagger}$}&{78.1$^{\dagger}$} & {\textbf{\bb{87.2}}$^{\dagger}$}& {\textbf{88.9}$^{\dagger}$}\\
&& iLPC (ICCV'21) & {\textbf{\bb{71.5}}$^{\dagger}$/\textbf{74.3}$^{\dagger}$} & {78.6$^{\dagger}$/81.1$^{\dagger}$}& {81.0$^{\dagger}$/83.3$^{\dagger}$}&{78.7$^{\dagger}$/\textbf{\bb{86.4}}$^{\dagger}$} &{85.6$^{\dagger}$/90.1$^{\dagger}$}& {86.9$^{\dagger}$/90.8$^{\dagger}$}&{\textbf{\bb{76.6}}$^{\dagger}$/\textbf{81.6}$^{\dagger}$} & {83.2$^{\dagger}$/\textbf{\bb{87.5}}$^{\dagger}$}& {85.1$^{\dagger}$/89.0$^{\dagger}$}&{80.8$^{\dagger}$} & {85.4$^{\dagger}$}& {86.5$^{\dagger}$}\\
&& BAVAR (AISTATS'23) & {\textbf{74.1}/\textbf{\bb{74.2}}$^{\dagger}$} & {\textbf{82.6}$^{\dagger}$/\textbf{\bb{82.6}}$^{\dagger}$}& {\textbf{85.5}/\textbf{\bb{85.6}}$^{\dagger}$}&{\textbf{82.0}/85.7$^{\dagger}$} &{\textbf{89.2}$^{\dagger}$/89.8$^{\dagger}$}& {\textbf{90.7}/90.8$^{\dagger}$}&{\textbf{77.5}/80.6$^{\dagger}$} & {\textbf{84.8}$^{\dagger}$/87.2$^{\dagger}$}& {\textbf{\bb{87.5}}/89.3$^{\dagger}$}&{\textbf{81.8}$^{\dagger}$} & {87.0$^{\dagger}$}& {87.9$^{\dagger}$}\\
&& 
PUTM (Ours) 
& {70.9/73.8} & {\textbf{\bb{80.5}}/\textbf{82.9}}& {84.6/\textbf{85.7}}&{\textbf{\bb{78.9}}/\textbf{86.9}} & {\textbf{\bb{87.8}}/\textbf{91.3}}& {89.3/\textbf{\bb{92.2}}} &{75.0/\textbf{\bb{81.1}}} & {\textbf{\bb{83.6}}/\textbf{88.4}}& {86.0/\textbf{90.4}}&{\textbf{\bb{81.4}}} & {\textbf{87.3}}& {\textbf{\bb{88.6}}}\\
\cmidrule{2-15}
&\multirow{8}{*}{\rotatebox{90}{Balanced}}
& PT-MAP (ARXIV) & {\textbf{78.9}/\textbf{82.5}$^{\dagger}$} & {\textbf{86.3}$^{\dagger}$/\textbf{87.5}$^{\dagger}$}& {86.6/\textbf{88.6}$^{\dagger}$}&{\textbf{\bb{85.5}}/92.0$^{\dagger}$} & {\textbf{\bb{90.5}}$^{\dagger}$/\textbf{94.0}$^{\dagger}$}& {\textbf{\bb{91.3}}/\textbf{94.3}$^{\dagger}$}&{\textbf{84.6}/\textbf{88.1}$^{\dagger}$} & {\textbf{88.2}$^{\dagger}$/\textbf{91.8}$^{\dagger}$}& {\textbf{90.0}/\textbf{92.5}$^{\dagger}$}&{\textbf{87.6}$^{\dagger}$} & {\textbf{90.5}$^{\dagger}$} & {\textbf{90.9}$^{\dagger}$}\\
&& LaplacianShot (ICML'20) & {72.9/73.4$^{\dagger}$} & {79.0$^{\dagger}$/80.5$^{\dagger}$}& {83.8/84.2$^{\dagger}$}&{78.9/87.5$^{\dagger}$} & {86.7$^{\dagger}$/90.4$^{\dagger}$}& {88.8/92.3$^{\dagger}$}&{78.8/78.7$^{\dagger}$} & {84.6$^{\dagger}$/85.6$^{\dagger}$}& {87.3/87.0$^{\dagger}$}&{82.1$^{\dagger}$} & {86.3$^{\dagger}$}& {86.8$^{\dagger}$}\\
&& TIM (NeuralIPS'20) & {74.6/77.6$^{\dagger}$} & {85.5$^{\dagger}$/86.0$^{\dagger}$}& {85.9/88.0$^{\dagger}$}&{80.3/88.1$^{\dagger}$} & {89.8$^{\dagger}$/93.0$^{\dagger}$}& {90.5/\textbf{\bb{94.0}}$^{\dagger}$}&{80.3/83.8$^{\dagger}$} &{\textbf{\bb{87.7}}$^{\dagger}$/90.6$^{\dagger}$}& {88.9/92.1$^{\dagger}$} &{83.0$^{\dagger}$} & {89.5$^{\dagger}$}& {\textbf{\bb{90.5}}$^{\dagger}$} \\
&& BD-CSPN (ECCV'20) & {72.5/73.8$^{\dagger}$} & {79.2$^{\dagger}$/80.7$^{\dagger}$}& {83.7/85.3$^{\dagger}$}&{77.9/87.3$^{\dagger}$} & {86.9$^{\dagger}$/91.4$^{\dagger}$}& {88.9/92.4$^{\dagger}$}&{77.7/78.7$^{\dagger}$} & {84.5$^{\dagger}$/86.4$^{\dagger}$}& {87.4/88.3$^{\dagger}$}&{81.6$^{\dagger}$} & {87.4$^{\dagger}$}& {88.6$^{\dagger}$}\\
&& $\alpha$-TIM (NeuralIPS'2021) & {71.0/71.5$^{\dagger}$} & {76.8$^{\dagger}$/77.7$^{\dagger}$}& {82.4/83.3$^{\dagger}$}&{77.2$^{\dagger}$/84.9$^{\dagger}$} & {85.1$^{\dagger}$/90.2$^{\dagger}$}& {87.6$^{\dagger}$/92.1$^{\dagger}$}&{75.1$^{\dagger}$/78.9$^{\dagger}$} & {80.6$^{\dagger}$/84.4$^{\dagger}$}& {85.3$^{\dagger}$/88.2$^{\dagger}$} &{79.7$^{\dagger}$} & {85.7$^{\dagger}$}& {87.9$^{\dagger}$}\\
&& iLPC (ICCV'21) & {{75.5}$^{\dagger}$/77.8$^{\dagger}$} & {82.0$^{\dagger}$/83.9$^{\dagger}$}& {83.9$^{\dagger}$/85.7$^{\dagger}$}&{82.4$^{\dagger}$/89.6$^{\dagger}$} &{87.9$^{\dagger}$/92.3$^{\dagger}$}& {88.9$^{\dagger}$/93.0$^{\dagger}$}&{81.0$^{\dagger}$/85.5$^{\dagger}$} & {86.4$^{\dagger}$/89.8$^{\dagger}$}& {87.5$^{\dagger}$/90.9$^{\dagger}$} &{84.8$^{\dagger}$} & {88.6$^{\dagger}$}& {89.2$^{\dagger}$}\\
&& BAVAR (AISTATS'23) & {\textbf{\bb{78.5}}/78.5$^{\dagger}$} & {84.8$^{\dagger}$/84.9$^{\dagger}$}& {\textbf{87.4}/87.7$^{\dagger}$}&{\textbf{85.6}/89.0$^{\dagger}$} &{\textbf{90.8}$^{\dagger}$/91.7$^{\dagger}$}& {\textbf{91.4}/92.1$^{\dagger}$}&{81.5/84.7$^{\dagger}$} & {87.4$^{\dagger}$/89.3$^{\dagger}$}& {88.3/90.3$^{\dagger}$}&{85.0$^{\dagger}$} & {88.4$^{\dagger}$}& {88.9$^{\dagger}$}\\
&&
PUTM(Ours) & {78.4/\textbf{\bb{81.4}}} & {\textbf{\bb{84.9}}/\textbf{\bb{86.7}}}& {86.5/\textbf{\bb{88.1}}} &{84.9/\textbf{\bb{91.2}}} & {89.2/\textbf{\bb{93.3}}}& {90.1/93.7} &{\textbf{\bb{81.6}}/\textbf{\bb{86.8}}} & {86.9/\textbf{\bb{90.7}}}& {\textbf{\bb{88.6}}/\textbf{\bb{92.2}}} &{\textbf{\bb{86.4}}} & {\textbf{\bb{89.5}}}& {90.4} \\
\bottomrule
\end{tabular}
}
\end{table*}

\begin{figure*}[!t]
\centerline{\includegraphics[width=1.\linewidth,height=0.25\linewidth]{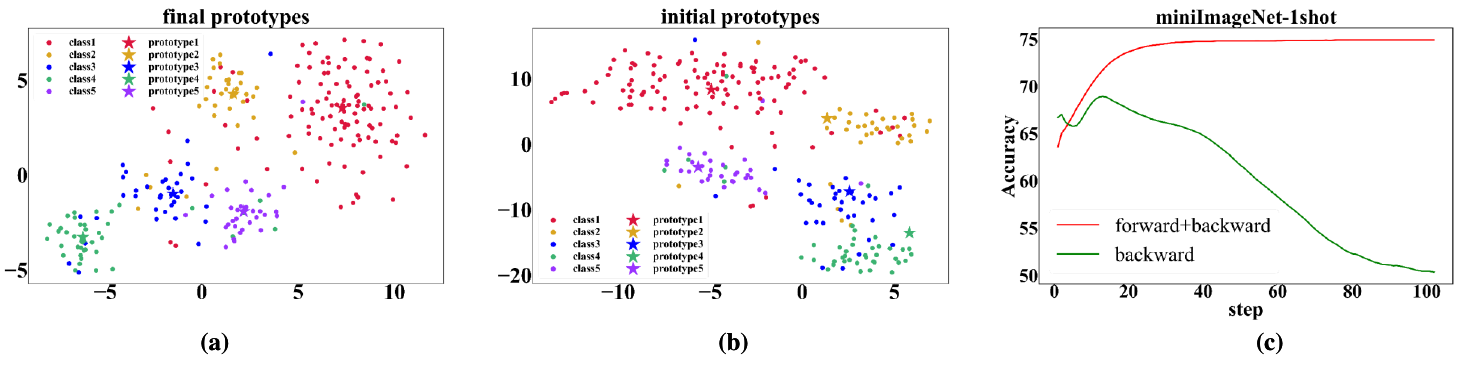}}
\caption{(a)-(b): The t-SNE \cite{van2008visualizing} visualizations before (middle) and after (left) calibration on a 5-way 1-shot miniImageNet task. Pentagrams are prototypes while solid dots are query features. Classes change and best view in colors. (c): Impact of forward and backward transport matrices in calibrating prototypes on miniImageNet over 3000 5-way 1-shot tasks.
}
\label{fig:tsne-fb}
\end{figure*}

\section{Experiments}
\subsection{Experimental Setup}
\label{esp}

\textbf{Baselines:} 
We mainly compare our model with some recently proposed State-Of-The-Art (SOTA) methods on one Nvidia RTX 3090 GPU
. For inductive methods, we adopt Baseline/Baseline(++) \cite{chen2019closer}, ProtoNet \cite{snell2017prototypical}, Simpleshot \cite{wang2019simpleshot}, and HOT \cite{guo2022adaptive} in inductive . For transductive methods, we adopt PT-MAP \cite{hu2021leveraging}, LaplacianShot \cite{ziko2020laplacian}, TIM\cite{boudiaf2020information}, BD-CSPN \cite{liu2020prototype}, $\alpha$-TIM \cite{veilleux2021realistic}, iLPC \cite{lazarou2021iterative}, and BAVAR \cite{hu2023adaptive}. 

\textbf{Datasets:} We evaluate our model on four benchmarks including miniImageNet \cite{ravi2017optimization}, tieredImageNet \cite{ren2018meta}, CUB \cite{wah2011caltech}, and CIFAR-FS \cite{bertinetto2018meta}. MiniImageNet is randomly chosen from ILSVRC-12 dataset \cite{deng2009imagenet} and it contains 100 classes with 600 images per class, the resolution of which is $84 \times 84 \times3$. Following previous works \cite{ravi2017optimization, veilleux2021realistic, hu2021leveraging, yang2021free}, the dataset is splitted into 64 base classes, 16 validation classes, and 20 novel classes. TieredImageNet \cite{ren2018meta} is also a subset of ILSVRC-12 dataset, which is larger than miniImageNet. It contains 608 classes sampled from a hierarchical category structure with around 1281 images per class. Following the previous work \cite{ren2018meta}, we split the dataset into 351 base classes, 97 validation classes, and 160 novel classes. The resolution is also $84 \times 84 \times 3$. CUB is a fine-grained dataset, containing 200 classes of birds with around 60 images per class. And the resolution is still $84 \times 84 \times 3$. We adopt 100 base classes, 50 validation classes, and 50 novel classes following \cite{chen2019closer}. CIFAR-FS consists of all 100 classes from CIFAR-100 \cite{xu2015empirical}, whose classes are randomly split into 64 base classes, 16 classes, and 20 novel classes. Each class contains 600 images of size $32 \times 32 \times 3$.

\textbf{Implementation details:} 
We use pre-trained backbones 
for extracting visual features of four benchmarks, as shown in Table. \ref{Table 1}.
WRN and RN18 are separately WideResNet-28-10 and ResNet-18 pre-trained with standard classification loss \cite{veilleux2021realistic}. WRN$\S$ is WideResNet-28-10 pre-trained with self-supervised loss \cite{mangla2020charting}.
Besides, we employ log transform, a channel-wise feature adaptation, introduced in Yang \etal \cite{yang2021free} to pre-process the pre-extracted features.
The distance function $\phiv$ is instantiated as a two-layers MLP activated with LeakyReLU whose architecture is $[d_f,128,64,1]$.
As for hyper-parameters, we mainly set $\beta=0.5$, $\alpha=0.1$, $\rho=0.2$.
The reported results are averaged over 3000 tasks.

\subsection{Main Results}
\textbf{Quantitative results:} 
The classification results on miniImageNet, tieredImageNet, CUB, and CIFAR-FS are reported in Table \ref{Table 1}, according to which five-fold observations can be drawn: 1) inductive methods are not bothered by class-imbalanced problem since they do not leverage transferable statistics of query samples; 2) most transductive methods still outperform the inductive peers in class-imbalanced setting, which implies that there exists a trade-off between extra statistics conveyed by class-imbalanced query samples and algorithm bias; 
3) results of features extracted by WRN$\S$ usually outperforms those by WRN or RN18, it indicates that self-supervised learning is helpful in obtaining generalized representations in class-imbalanced and class-balanced FSL applications;
4) most existed transductive inference methods suffer from performance cliff since their mismatch between data distribution and prior distribution. For example, TIM implicitly introduces uniform prior on query samples by using entropy regularization since minimizing entropy-based regularization is equivalent to minimizing the KL divergence between the predicted marginal distribution and uniform distribution. Hence, data distribution and model bias are unsuitable, leading to poor accuracy. Fortunately, such kind of mismatch can be eased by re-weighting in \cite{veilleux2021realistic}. On the contrary, PT-MAP holds a strong but unmodified uniformly distributed prior on its transport matrix. Therefore, results of TIM are robuster than those of PT-MAP;
5) our proposed method outperforms most of the transductive methods and achieves competitive results with some most recently proposed methods, such as $\alpha$-TIM, iLPC and BAVAR
. It indicates that prototype refinement benefits from the well optimized transport matrices when shot number is low. 
Moreover, compared with iLPC and BAVAR, on the one hand, our model is easy to implement. On the other hand, BAVAR heavily relies on prior of imbalanced ratio $\alphav$ in class-imbalanced case while iLPC depends on uniform prior in class-balanced case \cite{hu2023adaptive}, importantly however, our model is free of these priors, making it suitable for real-world applications when we have no idea on class distributions.

\begin{figure}[!t]
\centerline{\includegraphics[width=1.05\linewidth,height=0.35\linewidth]{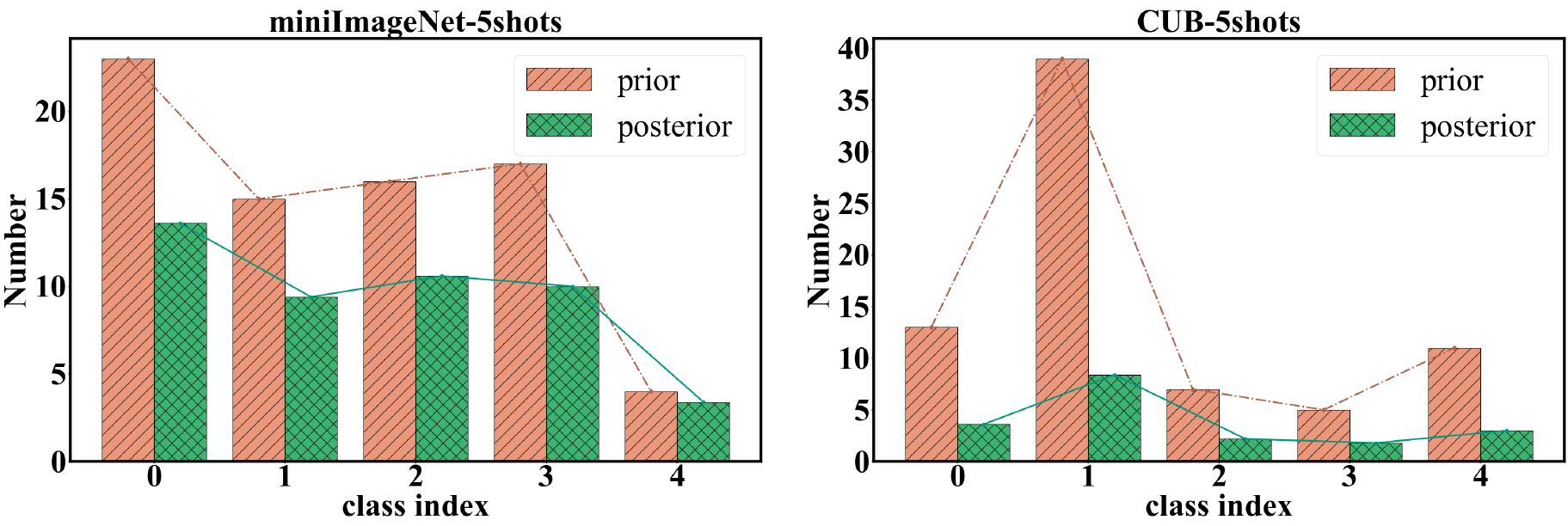}}
\caption{Case studies for 5-way 5-shots tasks on CUB \cite{wah2011caltech} (left) and miniImageNet (right). Comparisons between unnormalized prior distribution (true number of query samples counting for each class) and unnormalized posterior distribution (estimated number of query samples counting by the well optimized forward transport matrix for each class). 
}
\label{fig:navigators}
\end{figure}

\textbf{Qualitatively analysis:} To qualitatively illustrate the evolution of prototypes of our model, we map prototypes and query features of one task into a 2-D subspace with t-distributed Stochastic Neighbor Embedding (t-SNE) \cite{van2008visualizing}, it is a non-linear dimensional reduction technique well-suited for embedding high-dimensional data into a low-dimensional space. 
As we can see from Fig. \ref{fig:tsne-fb}, due to class-imbalanced issue, although prototypes can not match their corresponding class centers well at beginning, they become consistent with the ground truth centers after refinement using our proposed method, verifying the effectiveness of our PUTM in dealing with unknown and arbitrary prior situation.

\begin{figure}[!t]
\centerline{\includegraphics[width=1.05\linewidth,height=0.4\linewidth]{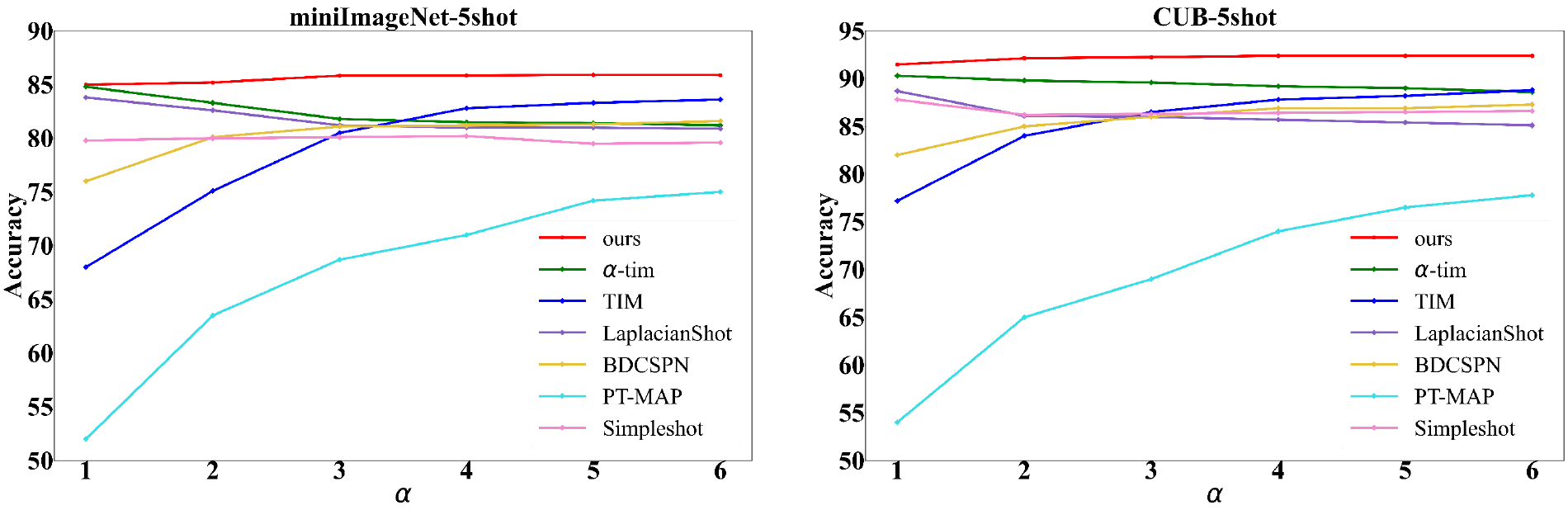}}
\caption{5-way 5-shot classification accuracy on miniImageNet and CUB of transductive methods versus imbalance level, lower $\alphav$ corresponds to severer class imbalance.
}
\label{fig:dynamic}
\end{figure}

\begin{table}[!t]
\centering
\caption{{End-to-end real-time efficiencies ($s$) of 5-way 1-shot on CUB with two backbones averaged on 100 tasks.}}
\label{Table 2}
\resizebox{1\hsize}{!}{
\begin{tabular}{ccccc}
\toprule 
\multicolumn{1}{c}{\multirow{2}*{Backbone}} & \multicolumn{1}{c}{$\alpha$ - TIM} & \multicolumn{1}{c}{iLPC} & \multicolumn{1}{c}{BAVAR}  & \multicolumn{1}{c}{Ours}\\
\multicolumn{1}{c}{ }  & \multicolumn{1}{c}{(NeurIPS'21)} & \multicolumn{1}{c}{(ICCV'21)} & \multicolumn{1}{c}{(AISTATS'23)} \\
\cmidrule{2-5}
RN18 & {5.6} & {4.6} & {3.4} & {2.9} \\
WRN & {6.2} & {4.7} & {3.6} & {3.7} \\
\bottomrule
\end{tabular}
}
\end{table}

\begin{figure*}[!t]
\centerline{\includegraphics[width=1\linewidth,height=0.4\linewidth]{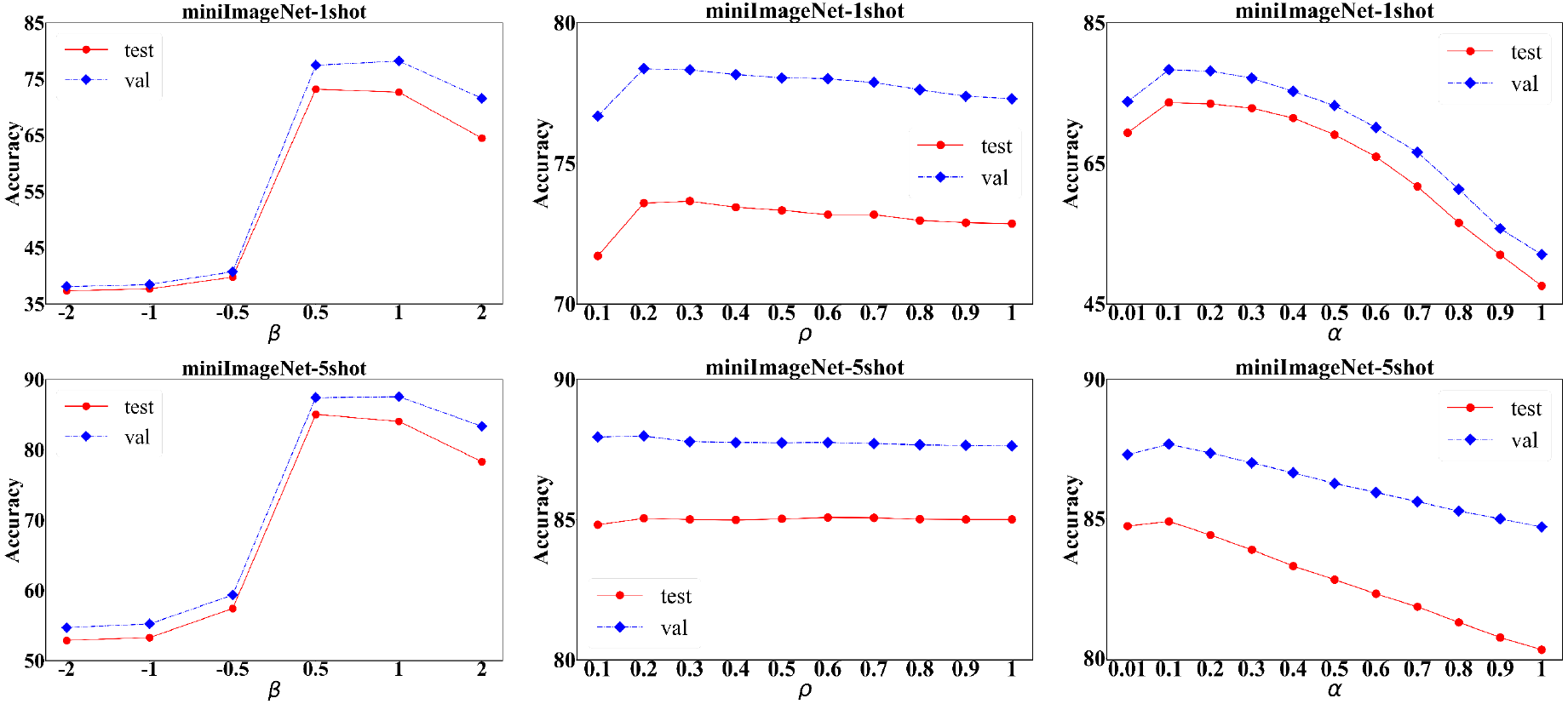}} 
\caption{Sensitivity analysis of our model to hyper-parameters including log transform $\beta$, CT coefficient $\rho$, and inertia $\alpha$. Results are obtained with WRN$\S$ and are reported on both validation and test sets. Best viewed in color.
}
\label{fig:ablation}
\end{figure*}

\subsection{Model Analysis}
\textbf{Forward transport V.S. backward transport:} 
As discussed in Sec. \ref{UST}, the forward and backward transport matrices put different priors of $\{a_i\}_{i=1}^N$ 
. Namely, we do not restrict uniform distribution in the forward transport and enable it to be adaptive to data-driven class distribution. On the contrary, prior in the backward transport is always assumed to be $\frac{1}{N}$. Therefore, we employ the forward transport for predictions to meet the class-imbalanced essence, the efficiency is illustrated in Fig. \ref{fig:navigators},
it depicts that the predicted distribution is consistent with the true distribution.
Moreover, we conduct an extra experiment to verify the opposite direction of the two transport matrices, results are illustrated in Fig. \ref{fig:tsne-fb}, where we separately use the forward transport and the backward transport to calibrate the prototypes over 3000 tasks. According to the results, we find that, from a statistical point of view, 
the backward transport does harm the efficient migration from the class-imbalanced query samples. 

\textbf{Robustness of imbalanced coefficients:} 
An important point in this work is that our model is robust for various class-imbalanced coefficients, which is vital and crucial for real-world applications since we usually do not know whether the few-shot tasks are balanced or not in advance, thus, we could not know what kind of methods are suitable for describing the samples distribution of tasks. Once they are unmatched, horrible performance degradation would happen, such as PT-MAP does.
We verify the robustness of our model with a wide range of imbalanced coefficients as shown in Fig. \ref{fig:dynamic}, where $\alphav$ changes from 1 to 6. 
The performance of most methods decreases as $\alphav$ decreasing because smaller $\alphav$ indicates more imbalanced distribution of query samples, thus making tasks harder. 
On the contrary, performances of LaplacianShot and $\alpha$-TIM drop obviously as $\alphav$ increases. 
Our model achieves consistent improvement and reasonable trend in a wide range of $\alphav$, we attribute this to the introduction of adaptive transport matrices, which enables PUTM to fit the class distributions precisely and flexibly.

\textbf{Sensitivity of hyper-parameters:}
To further verify the robustness of our model, we validate the sensitivity of PUTM to different hyper-parameters introduced in Sec. \ref{esp}, and results are reported in Fig. \ref{fig:ablation}. Generally speaking, the tendencies on validation and test sets are same. Besides there are three-fold observations: i) CT coefficient is less sensitive than log transform and inertia; ii) log transform controls the strength of channel-wise attention, which is proven to be important especially in few-shot case, thus it affects performance mostly; iii) inertia reflects the bottleneck of information flow from the aggregated prototypes to the refined prototypes. Bigger or smaller value of inertia is helpless for robust transfer since refinement of prototypes is associated with transport matrices. Namely, if the matrices are not well-optimized, 
a big inertia may cause algorithm to fall into a local optimum and vice versa.


\textbf{Real-time efficiency:}
To verify the efficiency of PUTM, we conduct parallel versions of $\alpha$-TIM, iLPC, and BAVAR based on their official implementations with the same machine. Results are reported in Table. \ref{Table 2}, as we can see that
our model consumes less time using RN18 and achieves competitive results using WRN. And we attribute this to the fact that
: i)
both $\alpha$-TIM and iLPC employ Sinkhorn algorithm to solve the Optimal Transport (OT) problem involving thousands of inner loop, and it is a kind of time consuming;
; ii) our EM-Solver might be more efficient for dealing with the class-imbalanced TFSL since it imposes potential capability for modeling such unknown distribution with CT theory.

\section{Conclusion and Future Works}
We propose a novel model called PUTM with EM-Solver based on CT theory for both class-imbalanced and class-balanced TFSL in this work according to an important observation that CT has a natural connection with unknown and arbitrary class prior problem. Specifically, PUTM exploits transferable statistics with CT for refining biased prototypes solved by EM to capture the unbiased statistics within imbalanced query samples. 
Extensive experiments on standard benchmarks verify the effectiveness of our model in both class-imbalanced and class-balanced few-shot applications.
In the future, we will explore the applicability of PUTM on more problem settings and applications, such as multi-domain few-shot classification and few-shot object detection.

\section{Acknowledgement}
We thanks Dandan Guo and He Zhao for their constructive comments on transductive few-shot learning, we thanks Zhisong Zhang for his very initial exploration on class-imbalanced few-shot image classification, we thanks Dongsheng Wang for his passion discussions on CT. Especially, we thanks Mingyuan Zhou's group for their elegant CT work.

{\small
\bibliographystyle{ieee_fullname}
\bibliography{egbib}
}

\end{document}